\pdfoutput=1

\documentclass[11pt]{article}

\usepackage[]{emnlp2021}

\usepackage{times}
\usepackage{latexsym}

\usepackage[T1]{fontenc}

\usepackage[utf8]{inputenc}

\usepackage{microtype}

\usepackage{color,array}
\usepackage{microtype}
\usepackage{multirow}
\usepackage{subcaption}
\usepackage{amsmath,amssymb}
\usepackage{graphicx}
\usepackage{comment}
\usepackage{graphicx}

\title{Enhancing Interpretable Clauses Semantically using Pretrained Word Representation}

\author{Rohan Kumar Yadav, Lei Jiao, Ole-Christoffer Granmo, and Morten Goodwin \\
  Centre for Artificial Intelligence Research \\ University of Agder \\
  4879, Grimstad, Norway \\

  \texttt{\{rohan.k.yadav, lei.jiao, ole.granmo, morten.goodwin\}@uia.no} \\}
\begin{document}
\maketitle

\begin{abstract}
Tsetlin Machine (TM) is an interpretable pattern recognition algorithm based on propositional logic, which has demonstrated competitive performance in many Natural Language Processing (NLP) tasks, including sentiment analysis, text classification, and Word Sense Disambiguation.  To obtain human-level interpretability, legacy TM employs Boolean input features such as bag-of-words (BOW). However, the BOW  representation makes it difficult to use any pre-trained information, for instance, word2vec and GloVe word representations. This restriction has constrained the performance of TM compared to deep neural networks (DNNs) in NLP. To reduce the performance gap, in this paper, we propose a novel way of using pre-trained word representations for TM. The approach significantly enhances the performance and interpretability of TM. We achieve this by extracting semantically related words from pre-trained word representations as input features to the TM. Our experiments show that the accuracy of the proposed approach is significantly higher than the previous BOW-based TM, reaching the level of DNN-based models.

\end{abstract}

\section{Introduction}
Tsetlin Machine (TM) is an explainable pattern recognition approach that solves complex classification problems using propositional formulas~\cite{Granmo2018TheTM}. 
Text-~\cite{Berge2019UsingTT}, numerical data-~\cite{10.1007/978-3-030-22999-3_49}, and image classification~\cite{granmo2019convolutional} are recent areas of application.  In Natural Language Processing~(NLP), TM has provided encouraging trade-offs between accuracy and interpretability for various tasks. These include Sentiment Analysis~(SA)~\cite{yadav2021sentiment,rupsa2020sentiment}, Word Sense Disambiguation~(WSD)~\cite{icaart21rohan}, and novelty detection~\cite{icaart21bimal}. Because TM NLP models employ bag-of-words~(BOW) that treat each word as independent features, it is easy for humans to interpret them. The models can be interpreted simply by inspecting the words that take part in the conjunctive clauses. However, using a simple BOW makes it challenging to attain the same accuracy level as deep neural network (DNN) based models.

\par A key advantage of DNN models is distributed representation of words in a vector space. By using a single-layer neural network, Mikolov et al. introduced such a representation, allowing for relating words based on the inner product between word vectors \cite{Mikolov2013DistributedRO}. One of the popular methods is skip-gram, an approach that learns word representations by predicting the context surrounding a word within a given window length. However, skip-gram has the disadvantage of not considering the co-occurrence statistics of the corpus. Later, Pennington et al. developed \emph{GloVe} -- a model that combines the advantages of local window-based methods and global matrix factorization~\cite{Pennington2014GloVeGV}. 
The foundation for the above vector representation of words is the distributional hypothesis that states that ``the word that occurs in the same contexts tend to have similar meanings'' \cite{doi:10.1080/00437956.1954.11659520}. This means that in addition to forming a rich high-dimensional representation of words, words that are closer to each other in vector space tend to represent similar meaning. As such, vector representations have been used to enhance for instance information retrieval \cite{manning_raghavan_schutze_2008}, 
name entity recognition \cite{turian}, 
and parsing \cite{socher}. 

\par The state of the art in DNN-based NLP has been advanced by incorporating various pre-trained word representations such as GloVe~ \cite{Pennington2014GloVeGV}, word2vec~\cite{Mikolov2013DistributedRO}, and fasttext~\cite{bojanowski2017enriching}. Indeed, building semantic representations of the words has been demonstrated to be a vital factor for improved performance. Most DNN-based models utilize the pre-trained word representations to initialize their word embeddings. This provides them with additional semantic information that goes beyond  a traditional BOW.

However, in the case of TM, such word representations cannot be directly employed because they consist of floating-point numbers. First, these numbers must be converted into Boolean form for TM to use, which may result in information loss. Secondly, replacing the straightforward BOW of a TM with a large number of floating-point numbers in fine-grained Boolean form would impede interpretability. In this paper, we propose a novel pre-processing technique that evades the above challenges entirely by extracting additional features for the BOW. The additional features are found using the pre-trained distributed word representations to identify words that enrich the BOW, based on cosine similarity. In this way, TM can use the information from word representations for increasing performance, and at the same time retaining the interpretability of the model.

\par The rest of the paper is organised as follows. We summarize related work 
in Section~\ref{RW}. The proposed semantic feature extraction for TM is then explained in Section~\ref{model}. In Section~\ref{TM}, we present the  TM architecture employing the proposed feature extension. We provide extensive experiment results in Section~\ref{exp}, demonstrating the benefits of our approach, before concluding the paper in Section~\ref{conc}.

\section{Related Work}\label{RW}
Conventional text classification usually focuses on feature engineering and classification algorithms. One of the most popular feature engineering approaches is the derivation of BOW features. Several complex variants of BOW have been designed such as $n$-grams~\cite{wang2012} and entities in ontologies~\cite{ChenthamarakshanMSL11}. Apart from BOW approaches, Tang et al. demonstrated a new mechanism for feature engineering using a time series model for short text samples~\cite{Tang2020EnrichingFE}. There are also several techniques to convert text into a graph and sub-graph \cite{rousseau2015, Luo2017BridgingSA}. In general, none of the above methods adopt any pre-trained information, hence have inferior performance.

\par Deep learning-based text classification either depends on initializing models from pre-trained word representations, or on jointly learning both the word- and document level representations. Various studies report that incorporating such word representations, embedding the words, significantly enhances the accuracy of text classification~\cite{joulin2017bag,shen2018}. Another approach related to pre-trained word embedding is to aggregate unsupervised word embeddings into a document embedding, which is then fed to a classifier \cite{pmlrle14, jianPTE}. 

\par Despite being empowered with world knowledge through pre-trained information, DNNs such as BERT~\cite{devlinbert} and XLNet~\cite{NEURIPS2019_dc6a7e65} can be very hard to interpret. 
One interpretation approach is to use attention-based models,  relying on the weights they assign to the inputs. 
However, more careful studies reveal that attention weights in general do not provide a useful explanation \cite{Bai2020WhyIA, serrano-attention}. Researchers are thus increasingly shifting focus to other kinds of machine learning, with the TM being a recent approach considered to provide human-level interpretability~\cite{Berge2019UsingTT, Granmo2018TheTM, yadav2021sentiment}. 
It offers a very simple model consisting of multiple Tsetlin Automata (TAs) that select which features take part in the classification. However, despite promising performance, there is still a performance gap to the DNN models that utilize pre-trained word embedding. Yet, several TM studies demonstrate high degree of interpretability through simple rules, with a marginal loss in accuracy~\cite{yadav2021sentiment, icaart21rohan, rupsa2020sentiment}.

A significant reason for the performance gap between TM-based and state-of-the-art DNN-based NLP models is that TM operates on Boolean inputs, lacking a method for incorporating pre-trained word embeddings. Without pre-trained information, TMs must rely on labelled data available for supervised learning. On the other hand, incorporating high-dimensional Booleanized word embedding vectors directly into the TM would significantly reduce interpretability. In this paper, we address this intertwined challenge.  We propose a novel technique that boosts the TM BOW approach, enhancing the BOW with additional word features. The enhancement consists of using cosine similarity between GloVe word representations to obtain semantically related words. We thus distill information from the pre-trained word representations for utilization by the TM. To this end, we propose two methods of feature extension: (1) using the $k$ nearest words in embedding space and (2) using words within a given similarity threshold, measured as cosine angle ($\theta$).  By adopting the two methods, we aim to reduce the current performance gap between interpretable TM and black-box DNN, by achieving either higher or similar accuracy, relying on pre-trained word embedding.

\section{Boosting TM BOW with Semantically Related Words} \label{model}
Here, we introduce our novel method for boosting the BOW of TM with semantically related words. The method is based on comparing pre-trained word representations using cosine similarity, leveraging distributed word representation. There are various distributional representations of words available. These are obtained from different corpora, using various techniques, such as word2vec, GloVe, and fastText. We here use GloVe because of its general applicability.

\subsection{Input Feature Extraction from Distributed Word Representation}
Distributed word representation does not necessarily derive word similarity based on synonyms but based on the words that appear in the same context. As such, the representation is essential for NLP because it captures the semantics interconnecting words. Our approach utilizes this property to expand the range of features that we can use in an interpretable manner in TM.

\par Consider a full vocabulary $W$ of $m$ words, $W = [w_1, w_2, w_3 \ldots, w_m]$. Further consider a particular sentence that is represented as a Boolean BOW \(X = [x_1, x_2, x_3 , \ldots, x_m]\). In a Boolean BOW, each element $x_r$, \(r=1,2,3,\ldots, m\), refers to a specific word $w_r$ in the vocabulary $W$. The element $x_r$ takes the value $1$ if the corresponding word $w_r$ is present in the sentence and the value $0$ if the word is absent.  Assume that \(n\) words are present in the sentence, i.e., $n$ of the elements in $X$ are $1$-valued. Our strategy is to extract additional features from these by expanding them using cosine similarity. To this end, we use a GloVe embedding of each present word \(w_r, r \in \{z | x_z = 1, z=1,2,3\ldots,m\}\). The embedding for word $w_r$ is represented by vector \(w_r^{e} \in \Re^d\), where \(d\) is the dimensionality of the embedding (typically varying from $25$ to $300$).

\begin{figure}
    \centering
    \includegraphics[width=1\columnwidth]{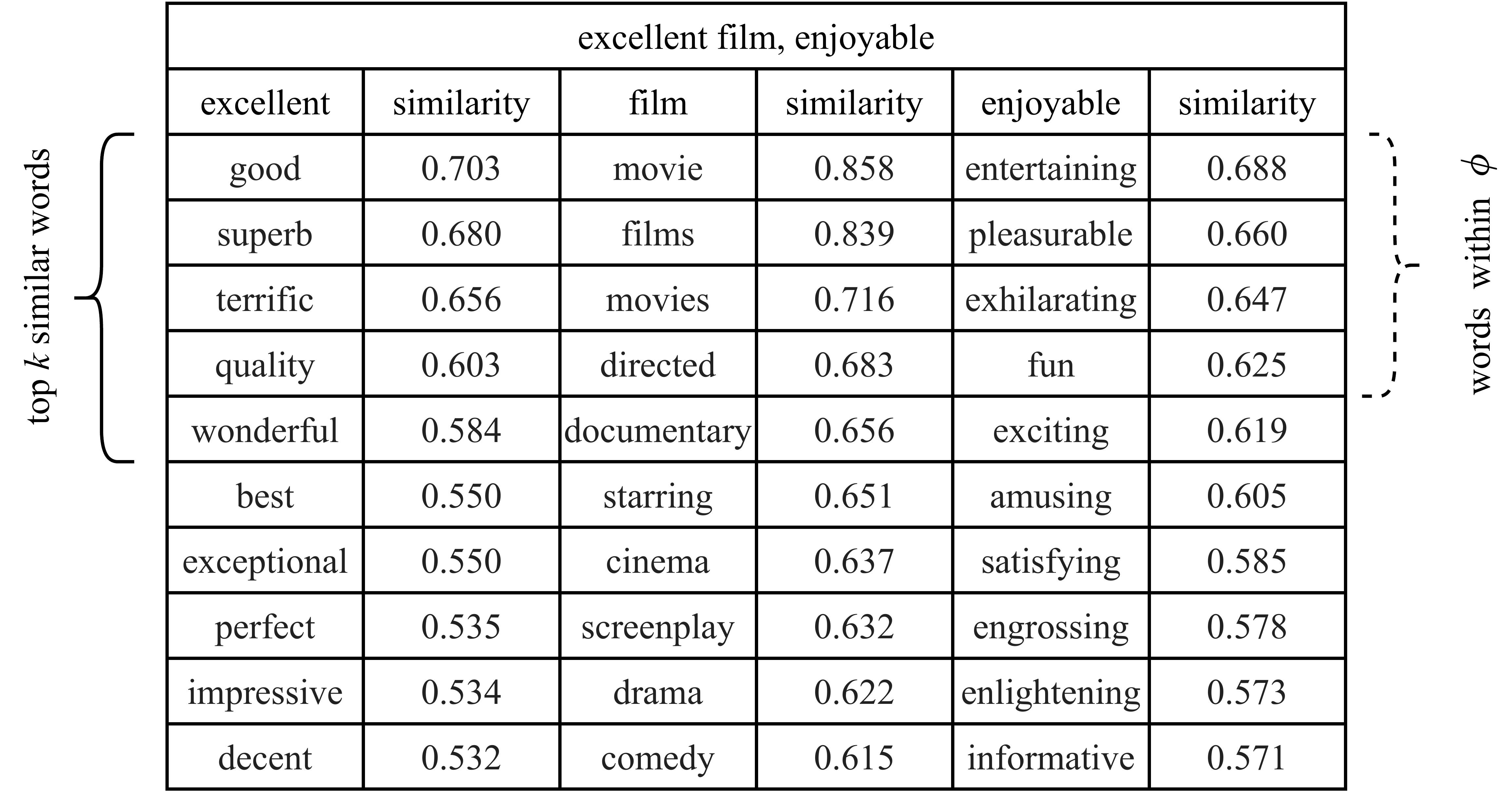}
    \caption{Similar words for an example ``excellent film, enjoyable'' using 300d GloVe word representation.}
    \label{fig2}
\end{figure}

\begin{figure}
    \centering
    \includegraphics[width=1\columnwidth]{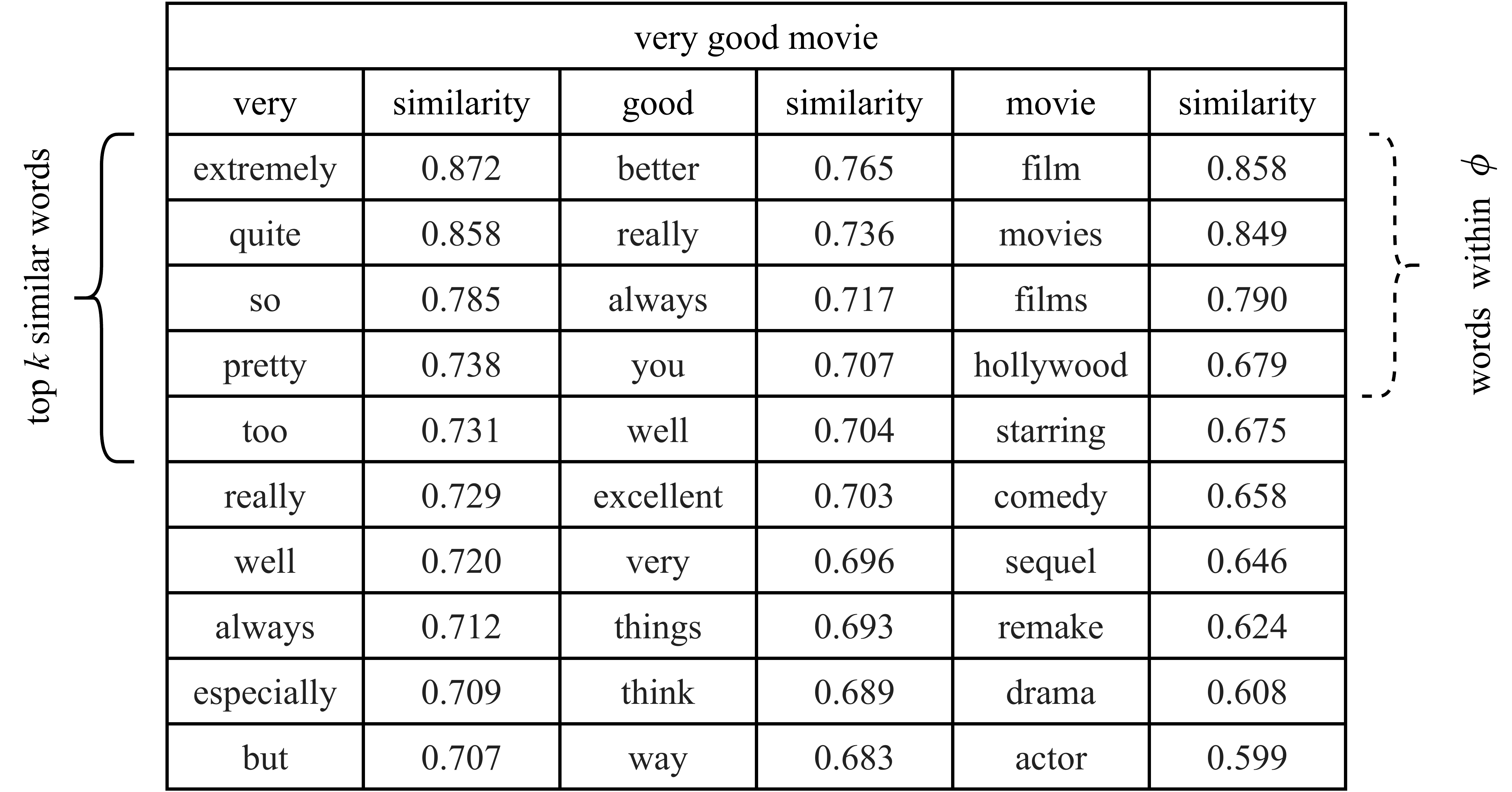}
    \caption{Similar words for an example ``very good movie'' using 300d GloVe word representation.}
    \label{fig3}
\end{figure}

We next introduce two selection techniques to expand upon each word: 
\begin{itemize}
    \item Select the top \(k\) most similar words,
    \item Select words up to a fixed similarity angle \(\cos(\theta) = \phi\).
\end{itemize}
\noindent For example, let us consider two contexts: ``very good movie'' and ``excellent film, enjoyable''.  Figs.~\ref{fig2} and \ref{fig3} list similar words showing the difference between top \(k\) words and words within angle \(\cos(\theta)\), i.e., \(\phi\). In what follows, we will explain how these words are found. 

\subsection{Similar Words based on Top \(k\) Nearest Words}
We first boost the Boolean BOW of the considered sentence by expanding $X$ with \((k-1) \times n\) semantically related words. That is, we add $k-1$ new words for each of the $n$ present words. We do this by identifying neighbouring words in the GloVe embedding space, using cosine similarity between the embedding vectors. 

 Consider the GloVe embedding vectors $W_G^e = [w_1^e, w_2^e, \ldots, w_m^e]$ of the full vocabulary $W$. For each word $w_r$ from the sentence considered, the cosine similarity to each word $w_t$, $t = 1,2,\ldots, m$, of the full vocabulary is given by Eq.~(\ref{eqn11}),
 
{\small
\begin{equation}\label{eqn11}
    \phi_r^t = \cos(w_r^{e}, w_t^e) = \frac{w_r^{e} \cdot w_t^e}{||w_r^{e}||\cdot ||w_t^e||}.
\end{equation}}
\noindent 
Clearly, \(\phi_r^t\) is the cosine similarity between \(w_r^{e}\) and \(w_t^e\).
By calculating the cosine similarity of $w_r$ to the words in the vocabulary, we obtain $m$ values: $\phi_r^t$, $t = 1,2,\ldots, m$. We arrange these values in a vector $\Phi_r$:
{\small\begin{equation}\label{eqn12}
   \Phi_r = [\phi_r^1, \phi_r^2, \ldots, \phi_r^m ].
\end{equation}}
The $k$ elements from $\Phi_r$ of largest value are then identified and their indices are stored in a new set~$A_r$.

Finally, a boosted BOW, referred to as $X_{mod}$, can be formed by assigning element $x_t$ value $1$ whenever one of the $A_r$ contains $t$, and $0$ otherwise:

{\small\begin{eqnarray}\label{eqn13}
X_{mod} = [x_1, x_2, x_3, \ldots, x_m], \\
    x_t = \begin{cases}
    1 & \exists r, t \in A_r\\
    0 & \nexists r, t \in A_r.
    \end{cases}\nonumber
\end{eqnarray}}
In addition, the vocabulary size for a particular task/dataset can be changed accordingly, which is usually less than $m$.
Note that implementation-wise, the GloVe library provides the top $k$ similar words of $w_r$ without considering the word $w_r$ itself, having similarity score $1$. Hence, using the GloVe library, $w_r$  must also be added to the boosted BOW.

\subsection{Similar Words within Cosine Angle Threshold}
Another approach to enrich the Boolean BOW of a sentence  is thresholding the cosine angle. This is different from the first technique because the number of additional words extracted will vary rather than being fixed. Whereas the first approach always produces $k-1$ new features for each given word, the cosine angle thresholding brings in all those words that are sufficiently similar. The cosine similarity threshold is given by $\phi = \cos(\theta)$,
where $\theta$ is the threshold for vector angle, while $\phi$ is the corresponding similarity score.

\par As per Eq. (\ref{eqn12}), we obtain $\Phi_r$, which consists of the similarity scores of the given word $w_r$ in comparison to the $m$ words in the vocabulary. Then, for each given word $w_r$, the indices of those scores $\phi_r^t$ that are greater than or equal to $\phi$ $(\phi_r^t\geq \phi)$  are stored in the set $A_r$. Similar to the first technique, the words in $W$ with the indices in $A_r$ are utilized to create $X_{mod}$ as given by Eq. (\ref{eqn13}).

\section{Tsetlin Machine-based Classification}\label{TM}
\subsection{Tsetlin Machine Architecture}

A TM is composed by TAs that operate with literals -- Boolean inputs and their negations -- to form conjunctions of literals (conjunctive clauses). A dedicated team of TAs builds each clause, with each input being associated with a pair of TAs. One TA controls the original Boolean input whereas the other TA controls its negation. The TA pair selects a combination of ``Include'' or ``Exclude'' actions, which decide the form of the literal to include or exclude in the clause.

\begin{figure}[h]
    \centering
    \includegraphics[width=1\columnwidth]{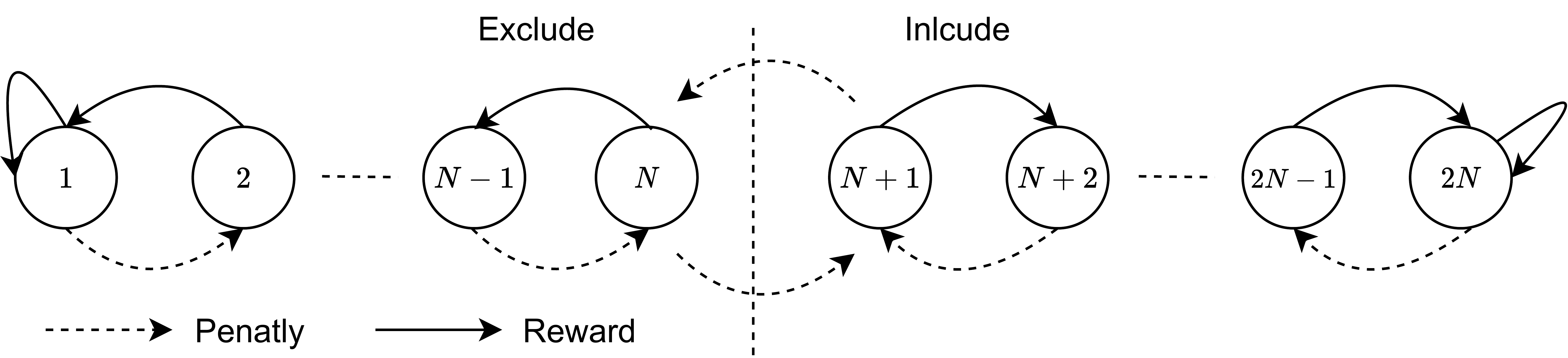}
    \caption{A TA with two actions: ``Include'' and ``Exclude''.}
    \label{fig4}
\end{figure}

Each TA decides upon an action according to its current state. There are \(N\) states per TA action, \(2N\) states in total. When a TA finds itself in states \(1\) to \(N\), it performs the ``Exclude'' action. When in states \(N+1\) to \(2N\), it performs the ``Include'' action. How the TA updates its state is shown in Fig.~\ref{fig4}. If it receives Reward, the TA moves to a deeper state thereby increasing its confidence in the current action. However, if it receives Penalty, it moves towards the centre, weakening the action. It may eventually jump over the middle decision boundary, to the other action. It is through this game of TAs that the TM shapes the clauses into frequent and discriminative patterns.

\par With respect to NLP, TM heavily relies on the Boolean BOW introduced earlier in the paper. We now make use of our proposed modified BOW \(X_{mod} = [x_1, x_2, x_3, \ldots, x_m]\). Let \(l\) be the number of clauses that represent each class of the TM, covering \(q\) classes altogether. Then, the overall pattern recognition problem is solved using \(l \times q\) clauses. Each clause \(C_i^j\), \(1\leq j \leq q\), \(1\leq i \leq l\) of the TM is given by $C_i^j = \left(\bigwedge \limits _{k \in I_i^j}{x_k} \right) \wedge \left(\bigwedge \limits _{k \in \bar I_i^j}{\neg x_k} \right)$, 
where \(I_i^j\) and \(\bar I_i^j\) are non-overlapping subsets of the input variable indices, \(I_j^i, \bar{I_j^i} \subseteq \{1, \ldots, m\}, I_j^i \cap \bar{I_j^i} = \emptyset\). The subsets decide which of the input variables take part in the clause, and whether they are negated or not. The indices of input variables in \(I_j^i\) represent the literals that are included as is, while the indices of input variables in \(\bar{I_j^i}\) correspond to the negated ones. Among the \(q\) clauses of each class, clauses with odd indexes are assigned positive polarity (+) whereas those with even indices are assigned negative polarity (-). The clauses with positive polarity vote for the target class and those with negative polarity vote against it. A summation operator aggregates the votes by subtracting the total number of negative votes from positive votes, as shown in~Eq. (\ref{eqn4}). 

{\small
\begin{equation}\label{eqn4}
\begin{tabular}{r}
    $f^j(X_{mod}) =\Sigma_{i=1,3,\ldots}^{l-1}C^j_i{(X_{mod})}-$ \\
    $\Sigma_{i=2,4,\ldots}^{l}C^j_i{(X_{mod})}$.
\end{tabular}
\end{equation}}

 For \(q\) number of classes, the final output \(y\) is given by the argmax operator to classify the input based on the highest sum of votes, $\hat{y} =\mathrm{argmax}_{j}\left( f^j(X_{mod}) \right)$. 

\subsection{Distributed Word Representation in TM}

\begin{figure}[h]
    \centering
    \includegraphics[width=1\columnwidth]{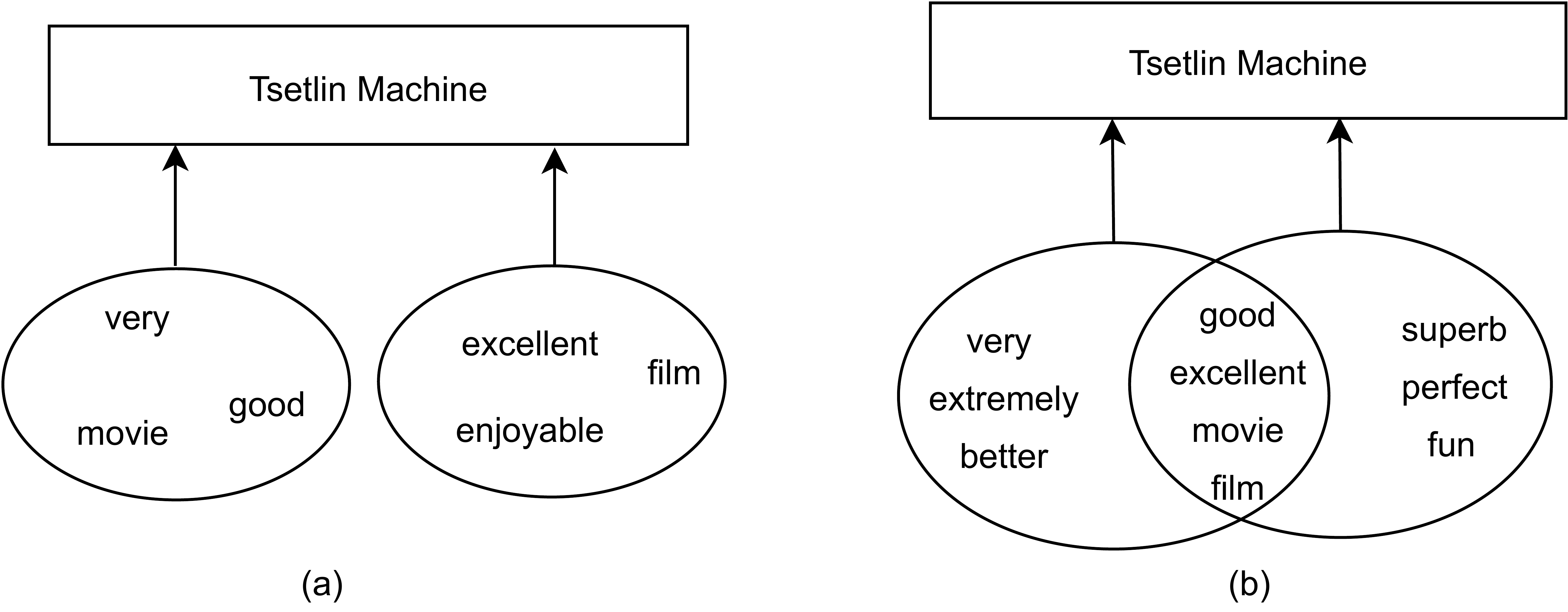}
    \caption{(a) BOW input representation without distributed word representation. (b) BOW input using similar words based on distributed word representation.}
    \label{fig5}
\end{figure}

Consider two contexts for sentiment classification: ``Very good movie'' and ``Excellent film, enjoyable''. Both contexts have different vocabularies but some of them are semantically related to each other. For example, ``good'' and ``excellent'' have similar semantics as well as ``film'' and ``movie''. Such semantics are not captured in the BOW-based input. However, as shown in Fig.~\ref{fig5}, adding words to the BOWs that are semantically related, as proposed in the previous section, makes distributed word representation available to the TM. 
\par
\begin{figure*}[h]
    \centering
    \includegraphics[width=0.8\textwidth]{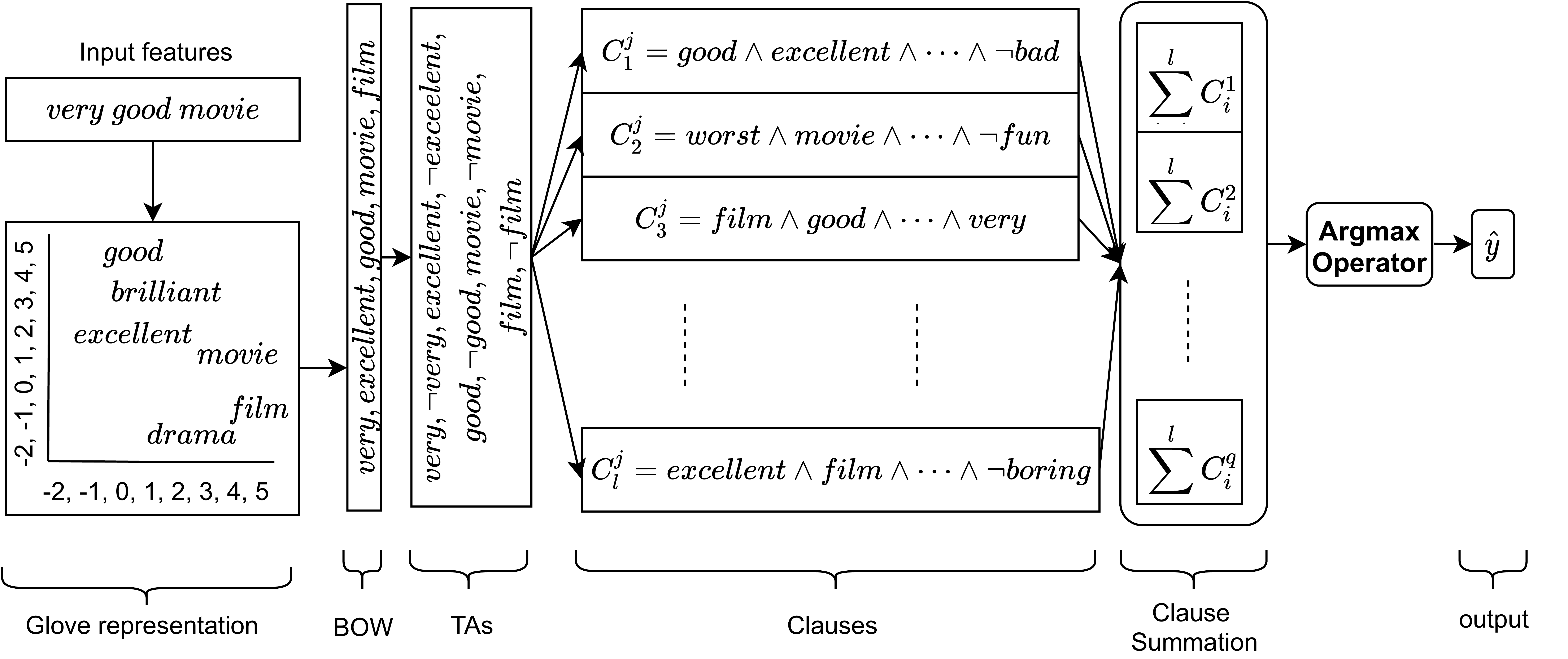}
    \caption{Architecture of TM using modified BOW based on word similarity.}
    \label{fig66}
\end{figure*}

    
 The resulting BOW-boosted TM architecture is shown in Fig. \ref{fig66}. Here each input feature is first expanded using the GloVe representation, adding semantically related words. Each feature is then transferred to its corresponding TAs, both in original and negated form. Each TA, in turn, decides whether to include or exclude its literal in the clause by taking part in a decentralized game. The actions of each TA is decided by its current state and updated by the the feedback it receives based on its action. As shown in the figure, the TA actions produce a collection of conjunctive clauses, joining the words into more complex linguistic patterns.
 
 There are two types of feedback that guides the TA learning. They are Type I feedback and Type~II feedback, detailed in~\cite{Granmo2018TheTM}. 
 Type I feedback is triggered when the ground truth label is \(1\), i.e., \(y = 1\). 
 The purpose of Type I feedback is to include more literals from the BOW to refine the clauses, or to trim them by removing literals. The balance between refinement and trimming is controlled by a parameter called specificity, \(s\). Type I feedback guides the clauses to provide true positive output, while simultaneously controlling over-fitting by producing frequent patterns. Conversely, Type II feedback is triggered in case of false positive output. Its main aim is to introduce zero-valued literals into clauses when they give false positive output. The purpose is to change them so that they correctly output zero later in the learning process. Based on these feedback types, each TA in a clause receives Reward, Penalty or Inaction. 
 The overall learning process is explained in detail by Yadav et al. in \cite{yadav2021sentiment}.

\section{Experiments and Results}\label{exp}
In this section, we evaluate our TM-based solution with the input features enhanced by distributed word representation. Here we use Glove pretrained word vector that is trained using CommonCrawl with the configuration of 42B tokens, 1.9M vocab, uncased, and 300d vectors. 
\subsection{Datasets}
    We have selected various types of datasets to investigate how broadly our method is applicable: R8 and R52 of Reuters, Movie Review (MR), and TREC-6. $\bullet$  \textbf{ Reuters} 21578 dataset include two subsets: R52 and R8 (all-terms version). R8 is divided into 8 sections while there are 52 categories in R52. 
    $\bullet$\textbf{MR} is a movie analysis dataset for binary sentiment classification with just one sentence per review \cite{10.3115/1219840.1219855}. 
    In this study, we used a training/test split from \cite{jianPTE}\footnote{https://github.com/mnqu/PTE/tree/master/data/mr.}.
    $\bullet$\textbf{TREC-6} is a question classification dataset \cite{li2002}. The task entails categorizing a query into six distinct categories (abbreviation, description, entity, human, location, numeric value). 


\subsection{TM Parameters}
A TM has three parameters that must be initialized before training a model: number of clauses $l$, voting target $T$, and specificity $s$. We configure these parameters as follows. For R8, we use 2,500 clauses, a threshold of 80, and specificity 9. The vocabulary size is 5,000. For R52, we employ  1,500 clauses, the voting target is 80, and specificity is 9. Here, we use a vocabulary of size 6,000. For MR, the number of clauses is 3,000, the voting target is 80, and specificity is 9, with a vocabulary of size 5,000. Finally, for TREC, we use 2,000 clause, a voting target of 80, and specificity 9, with vocabulary size 6,000. These parameters are kept static as we explore various $k$ and $\theta$ values for selecting similar words to facilitate comparison. 

\subsection{Performance When Using Top $k$ Nearest Neighbors}
Here, we demonstrate the performance on each of the datasets, exploring the effect of different $k$-values, i.e., $3$, $5$ and $10$. The performance of the proposed technique for selected datasets with various values of $k$ is compared in Table \ref{table55}. It can be seen that by using feature extension, performance is significantly enhanced. Both $k=3$ and $k = 5$ outperform the simple BOW ($k=0$). However, for this particular dataset, $k=10$ performs poorly because extending each word to its 10 nearest neighbors includes many unnecessary contexts that have no significant impact on the classification. In terms of accuracy, $k=5$ performs best for the R8 dataset. For the R52 dataset, the feature extension with $k=5$ and $k=10$ performs poorly compared to using $k=0$ and $k=3$. Here, $k=3$ is the best-performing parameter. The improvement obtained by moving from a simple BOW to a BOW enhanced with semantically similar features is obvious in the case of the R52 dataset. Similarly, in the case of the TREC dataset, the performance of simple BOW ($k=0$) is markedly outperformed by the feature extension techniques for all the tested $k$-values, with $k=5$ and $k=10$ being good candidates. The advantage of $k=10$ over $k=5$ is that $k=10$ reaches its peak accuracy  in an earlier epoch. Lastly, the performance of the MR is again clear that the feature extension technique outperforms the simple BOW ($k=0$) with a high margin.
\begin{table}
\centering
\begin{tabular}{lrlrlrlrlrl}
\hline \textbf{Parameters} & \textbf{R8} & \textbf{R52} & \textbf{MR} & \textbf{TREC} \\ \hline
k=0 & 96.16 & 84.62 & 75.14  & 88.05 \\
k=3 & 97.08 & 88.59 & 75.21  & 88.72 \\
k=5 & 96.80 & 70.60 & 76.06  & 89.16 \\
k=10 & 87.44 & 66.94 & 77.51  & 89.82 \\
\hline  
\end{tabular}
\caption{\label{table55} Comparison of feature extended TM with several parameters for $k$.}
\end{table}

\subsection{Performance When Using Neighbors Within a Similarity Threshold}
This section demonstrates the performance of our BOW enhancement approach  when using various similarity thresholds $\phi$ for feature extension. Here, $\phi$ refers to the cosine similarity between a word in the BOW and a target word from the overall vocabulary. Again, similarity is measured in the GloVe embedding  space as  the cosine of the angle $\theta$ between the embedding vectors compared, $\cos(\theta)$. For $\phi$, we here explore the  values $0.5$, $0.6$, $0.7$, $0.8$, and $0.9$, whose corresponding angles are $60^{\circ}$, $53.13^{\circ}$, $45.57^{\circ}$, $36.86^{\circ}$, and $25.84^{\circ}$, respectively. The performance of the various $\phi$-values for the selected dataset is shown in Table \ref{table66}. For R8 dataset, feature extension using $\phi=0.7$, $\phi=0.8$, and $\phi=0.9$ outperforms the simple BOW ($\phi=0$) where $\phi=0.7$ being the best. In case of the R52 dataset, all of the investigated $\phi$-values outperform the simple BOW ($\phi=0$) where $\phi=0.5$ and $\phi=0.8$ performs the best. Similar trend is observed in case of TREC and MR dataset where feature extension outperforms the simple BOW.

\par In most of the cases, however, a too strict similarity threshold $\phi$ tends to reduce performance because fewer features are added to the BOW. Even though using a looser similarity score thresholds also introduces unnecessary features, these do not seem to impact the formation of accurate clauses. Overall, our experiments show that using $\phi$-values from $0.5$ to $0.7$ peaks performance.

\begin{table}
\centering
\begin{tabular}{lrlrlrlrlrl}
\hline \textbf{Parameters} & \textbf{R8} & \textbf{R52} & \textbf{MR} & \textbf{TREC} \\ \hline
$\phi=0$ & 96.16 & 84.62 & 75.14  & 88.05 \\
$\phi=0.5$ & 88.08 & 89.14 & 73.24  & 90.04 \\
$\phi=0.6$ & 90.86 & 88.05 & 74.34  & 87.83 \\
$\phi=0.7$ & 96.53 & 88.51 & 76.55  & 89.38 \\
$\phi=0.8$ & 96.25 & 88.94 & 75.12  & 88.27 \\
$\phi=0.9$ & 96.39 & 87.50 & 74.59  & 87.39 \\
\hline
\end{tabular}
\caption{\label{table66} Comparison of feature extended TM with several parameters for $\phi$.}
\end{table}

\subsection{Comparison with Baselines}
We here compare our proposed model with selected text classification- and embedding methods. We have selected representative techniques from various main approaches, both those that leverage similar kinds of pre-trained word embedding and those that only use BOW. The selected baselines are:
    $\bullet$\textbf{TF-IDF+LR}: This is a bag-of-words model employing Term Frequency-Inverse Document Frequency (TF-IDF) weighting. Logistic Regression is used as a softmax classifier.
    $\bullet$\textbf{CNN}: The CNN-baselines cover both  initialization with random word embedding (CNN-rand) as well as initialization with pretrained word embedding (CNN-non-static) \cite{kim2014}.
    $\bullet$ \textbf{LSTM}: The LSTM model that we employ here is from \cite{Liu2016}, representing the entire text using the last hidden state. We tested this model with and without pre-trained word embeddings.
    $\bullet$ \textbf{Bi-LSTM}: Bi-directional LSTMs are widely used for text classification. We compare our model with Bi-LSTM fed with pre-trained word embeddings.
    $\bullet$\textbf{PV-DBOW}: PV-DBOW is a paragraph vector model where the word order is ignored. Logistic Regression is used as a softmax classifier \cite{pmlrle14}.
    $\bullet$ \textbf{PV-DM}: PV-DM is also a paragraph vector model, however with word ordering taken into account. Logistic Regression is used as a softmax classifier \cite{pmlrle14}.
     $\bullet$\textbf{fastText}: This baseline is a simple text classification technique that uses the average of the word embeddings provided by fastText as document embedding. The embedding is then fed to a linear classifier \cite{joulin2017bag}. We evaluate both the use of uni-grams and bigrams.
    $\bullet$ \textbf{SWEM} : SWEM applies simple pooling techniques over the word embeddings to obtain a document embedding \cite{shen2018baseline}.
     $\bullet$\textbf{Graph-CNN-C}: A graph CNN model uses convolutions over a word embedding similarity graph \cite{NIPS2016_04df4d43}, employing a Chebyshev filter.
    $\bullet$\textbf{\(S^2GC\)}: This technique uses a modified Markov Diffusion Kernel to derive a variant of Graph Convolutional Network (GCN) \cite{zhu2021simple}.
     $\bullet$\textbf{LguidedLearn}: It is a label-guided learning framework for text classification. This technique is applied to BERT as well \cite{Liu2020LabelguidedLF}, which we use for comparison purposes here.
    $\bullet$\textbf{Feature Projection (FP)}: It is a novel approach to improve representation learning through feature projection. Existing features are projected into an orthogonal space~\cite{qin-etal-2020-feature}.

\par From Table \ref{table33}, we observe that the TM approaches that employ either of our feature extension techniques outperform several word embedding-based Logistic Regression approaches, such as PV-DBOW, PV-DM, and fastText. Similarly, the legacy TM outperforms sophisticated models like CNN and LSTM based on randomly initialized word embedding. Still, the legacy TM falls of other models when they are initialized by pre-trained word embeddings. By boosting the Boolean BOW with semantically similar features using our proposed technique, however, TM outperforms LSTM (pretrain) on the R8 dataset and performs similarly on R52 and MR. In addition to this, our proposed approach achieves quite similar performance compared to BERT, even though BERT has been pre-trained on a huge text corpus. However, it falls slightly short of sophisticated fine-tuned models like Lguided-BERT-1 and Lguided-BERT-3. Overall, our results show that our proposed feature extension technique for TMs significantly enhances accuracy, reaching state of the art accuracy. Importantly, this accuracy enhancement does not come at the cost of reduced interpretability, unlike DNNs, which we discuss below. The state of the art for the TREC dataset is different from the other three datasets, hence we report results  separately in Table~\ref{table44}. These results clearly show that although the basic TM model does not outperform the recent DNN- and transformer-based models, the feature-boosted TM outperforms all of those models except understandably BAE:BERT \cite{garg2020bae}.

\begin{table}
\centering
\resizebox{1\columnwidth}{!}{
\begin{tabular}{clclclc}
\hline \textbf{Model} & \textbf{R8} & \textbf{R52} & \textbf{MR} \\ \hline
TF-IDF+LR & 93.74 & 86.95 & 74.59 \\
CNN-rand  & 94.02 & 85.37 & 74.98 \\
CNN-non-static & 95.71 & 87.59 & 77.75 \\
LSTM & 93.68 & 85.54 & 75.06 \\
LSTM (pretrain) & 96.09 & 90.48 & 77.33 \\
Bi-LSTM & 96.31 & 90.54 & 77.68 \\
PV-DBOW & 85.87 & 78.29 & 61.09 \\
PV-DM & 52.07 & 44.92 & 59.47 \\
fastText & 96.13 & 92.81 & 75.14 \\
fastText (bigrams)  & 94.74 & 90.99 & 76.24 \\
SWEM & 95.32 & 92.94 & 76.65  \\
LEAM & 93.31 & 91.84 & 76.95 \\
Graph-CNN-C & 96.99 & 92.74 & 77.22 \\
\(S^2GC\) & 97.40 & 94.50 & 76.70 \\
BERT & 96.02 & 89.66 & 79.24 \\
Lguided-BERT-1 & 97.49 & 94.26 & 81.03 \\
Lguided-BERT-3 & 98.28 & 94.32 & 81.06 \\
TM & 96.16\(\pm\)  1.52 & 84.62\(\pm\)  1.8 & 75.14\(\pm\)  1.2 \\
TM with $k$ & 97.50\(\pm\)  1.12 & 88.59\(\pm\)  1.2 & 77.51\(\pm\)  0.6 \\
TM with $\phi$ & 96.39\(\pm\)  1.0 & 89.14\(\pm\)  1.5 & 76.55\(\pm\)  0.9 \\
\hline
\end{tabular}
}
\caption{\label{table33} Comparison of feature extended TM with the state of the art for R8, R52 and MR. Reported accuracy of TM is the mean of last 50 epochs of 5 independent experiments with their standard deviation.}
\end{table}

\begin{table}[h]
\centering
\resizebox{1\columnwidth}{!}{
\begin{tabular}{lc}
\hline
\textbf{Model} & \textbf{TREC}\\
\hline
LSTM & 87.19 \\
FP+LSTM & 88.83 \\
Transformer & 87.33 \\ 
FP+Transformer & 89.51 \\ 
BAE: BERT & 97.6\\
TM \cite{Drago2021QuestionCU} & 87.20 \\
TM & 88.05\(\pm\)  1.52  \\ 
TM with $k$ & 89.82\(\pm\)  1.18  \\ 
TM with $\phi$ & 90.04\(\pm\)  0.94  \\\hline
\end{tabular}
}
\caption{Comparison of feature extended TM with the state of the art for TREC. Reported accuracy of TM is the mean of last 50 epochs of 5 independent experiments with their standard deviation.}\label{table44}
\end{table}

\subsection{Interpretation}
\begin{figure}[h]
    \centering
    \includegraphics[width=0.9\columnwidth]{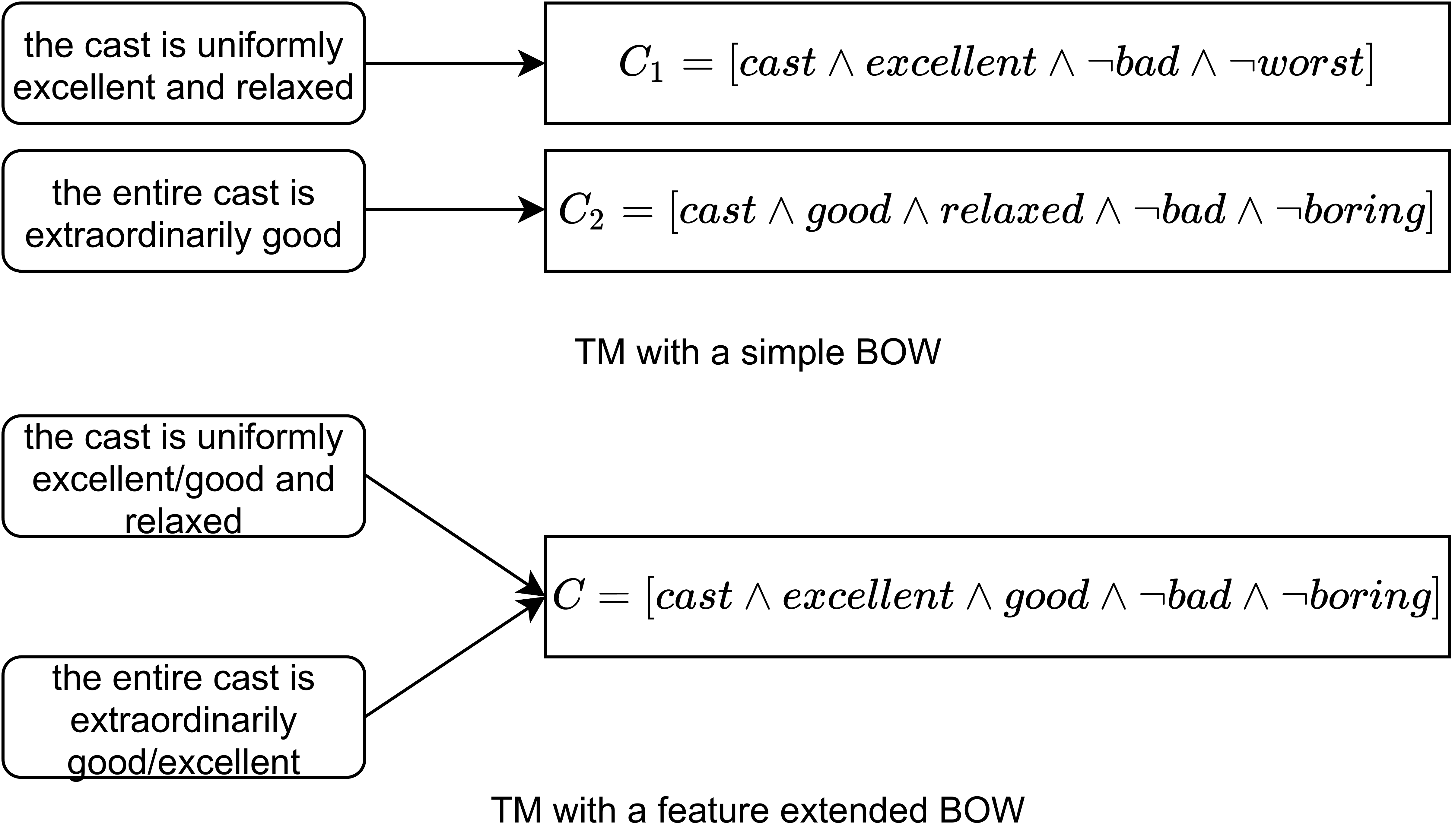}
    \caption{Clause learning semantic for multiple examples compared to simple BOW based TM.}
    \label{fig6}
\end{figure}

The proposed feature extension-based TM does not only impact accuracy. Perhaps surprisingly, our proposed technique also simplify the clauses that the TM produces, making them more meaningful in a semantic sense. To demonstrate this property, let us consider two samples from the MR dataset:  $S_1$=``the cast is uniformly excellent and relaxed'' and $S_2$=``the entire cast is extraordinarily good''. Let the vocabulary, in this case, be [cast, excellent, relaxed, extraordinarily, good, bad, boring, worst] as shown in Fig. \ref{fig6}.

As we can see, that the TM initialized with normal BOW uses two separate clauses to represent two examples. However, augmenting feature on TM uses only one clause that learns the semantic for multiple examples.This indeed makes interpretation of TM more powerful and meaningful as compared to simple BOW based TM.

\section{Conclusion}\label{conc}
In this paper, we aimed to enhance the performance of Tsetlin Machines (TMs) by introducing a novel way to exploit distributed feature representation for TMs. Given that a TM relies on Bag-of-words (BOW), it is not possible to introduce pre-trained word representation into a TM directly, without sacrificing the interpretability of the model. To address this intertwined challenge, we extended each word feature by using cosine similarity on the distributed word representation. We proposed two techniques for feature extension: (1) using the $k$ nearest words in embedding space and (2) including words within a given cosine angle ($\theta$). Through this enhancement, the TM BOW can be boosted with pre-trained world knowledge in a simple yet effective way. Our experiment results showed that the enhanced TM not only achieve competitive accuracy compared to state of the art, but also outperform some of the sophisticated deep neural network (DNN) models. In addition, our BOW boosting  also improved the interpretability of the model by increasing the scope of each clause, semantically relating more samples. We thus believe that our proposed approach significantly enhance the TM in the accuracy/interpretability continuum, establishing a new standard in the field of explainable NLP.

\bibliography{anthology,custom}

\begin{thebibliography}{39}
\expandafter\ifx\csname natexlab\endcsname\relax\def\natexlab#1{#1}\fi

\bibitem[{Abeyrathna et~al.(2019)Abeyrathna, Granmo, Zhang, and
  Goodwin}]{10.1007/978-3-030-22999-3_49}
Kuruge~Darshana Abeyrathna, Ole-Christoffer Granmo, Xuan Zhang, and Morten
  Goodwin. 2019.
\newblock A scheme for continuous input to the {T}setlin machine with
  applications to forecasting disease outbreaks.
\newblock In \emph{Advances and Trends in Artificial Intelligence. From Theory
  to Practice}, pages 564--578. Springer International Publishing.

\bibitem[{Bai et~al.(2020)Bai, Liang, Zhang, Li, Bai, and Wang}]{Bai2020WhyIA}
Bing Bai, J.~Liang, Guanhua Zhang, Hao Li, Kun Bai, and F.~Wang. 2020.
\newblock Why is attention not so interpretable.
\newblock \emph{arXiv: Machine Learning}.

\bibitem[{Berge et~al.(2019)Berge, Granmo, Tveit, Goodwin, Jiao, and
  Matheussen}]{Berge2019UsingTT}
Geir~Thore Berge, Ole-Christoffer Granmo, Tor~Oddbj{\o}rn Tveit, Morten
  Goodwin, Lei Jiao, and Bernt~Viggo Matheussen. 2019.
\newblock Using the tsetlin machine to learn human-interpretable rules for
  high-accuracy text categorization with medical applications.
\newblock \emph{IEEE Access}, 7:115134--115146.

\bibitem[{Bhattarai. et~al.(2021)Bhattarai., Granmo., and
  Jiao.}]{icaart21bimal}
Bimal Bhattarai., Ole{-}Christoffer Granmo., and Lei Jiao. 2021.
\newblock Measuring the novelty of natural language text using the conjunctive
  clauses of a tsetlin machine text classifier.
\newblock In \emph{Proceedings of the 13th International Conference on Agents
  and Artificial Intelligence - Volume 2: ICAART,}, pages 410--417. INSTICC,
  SciTePress.

\bibitem[{Bojanowski et~al.(2017)Bojanowski, Grave, Joulin, and
  Mikolov}]{bojanowski2017enriching}
Piotr Bojanowski, Edouard Grave, Armand Joulin, and Tomas Mikolov. 2017.
\newblock Enriching word vectors with subword information.
\newblock \emph{Transactions of the Association for Computational Linguistics},
  5:135--146.

\bibitem[{Chenthamarakshan et~al.(2011)Chenthamarakshan, Melville, Sindhwani,
  and Lawrence}]{ChenthamarakshanMSL11}
Vijil Chenthamarakshan, Prem Melville, Vikas Sindhwani, and Richard~D.
  Lawrence. 2011.
\newblock Concept labeling: Building text classifiers with minimal supervision.
\newblock In \emph{IJCAI}, pages 1225--1230.

\bibitem[{Defferrard et~al.(2016)Defferrard, Bresson, and
  Vandergheynst}]{NIPS2016_04df4d43}
Micha\"{e}l Defferrard, Xavier Bresson, and Pierre Vandergheynst. 2016.
\newblock Convolutional neural networks on graphs with fast localized spectral
  filtering.
\newblock In \emph{Advances in Neural Information Processing Systems},
  volume~29. Curran Associates, Inc.

\bibitem[{Devlin et~al.(2019)Devlin, Chang, Lee, and Toutanova}]{devlinbert}
Jacob Devlin, Ming-Wei Chang, Kenton Lee, and Kristina Toutanova. 2019.
\newblock {BERT}: Pre-training of deep bidirectional transformers for language
  understanding.
\newblock In \emph{ACL: Human Language Technologies, Volume 1 (Long and Short
  Papers)}, pages 4171--4186, Minneapolis, Minnesota. ACL.

\bibitem[{Dragoș et~al.(2021)Dragoș, Nicolae, and
  dragosnicolae}]{Drago2021QuestionCU}
Dragoș, Constantin Nicolae, and dragosnicolae. 2021.
\newblock Question classification using interpretable tsetlin machine.
\newblock In \emph{International Workshop of Machine Reasoning}. ACM
  International Conference on Web Search and Data Mining.

\bibitem[{Garg and Ramakrishnan(2020)}]{garg2020bae}
Siddhant Garg and Goutham Ramakrishnan. 2020.
\newblock \href {http://arxiv.org/abs/2004.01970} {Bae: Bert-based adversarial
  examples for text classification}.

\bibitem[{Granmo(2018)}]{Granmo2018TheTM}
Ole-Christoffer Granmo. 2018.
\newblock The tsetlin machine - a game theoretic bandit driven approach to
  optimal pattern recognition with propositional logic.
\newblock \emph{ArXiv}, abs/1804.01508.

\bibitem[{Granmo et~al.(2019)Granmo, Glimsdal, Jiao, Goodwin, Omlin, and
  Berge}]{granmo2019convolutional}
Ole-Christoffer Granmo, Sondre Glimsdal, Lei Jiao, Morten Goodwin, Christian~W.
  Omlin, and Geir~Thore Berge. 2019.
\newblock The convolutional tsetlin machine.
\newblock \emph{arXiv}, 1905.09688.

\bibitem[{Harris(1954)}]{doi:10.1080/00437956.1954.11659520}
Zellig~S. Harris. 1954.
\newblock Distributional structure.
\newblock \emph{WORD}, 10(2-3):146--162.

\bibitem[{Joulin et~al.(2017)Joulin, Grave, Bojanowski, and
  Mikolov}]{joulin2017bag}
Armand Joulin, Edouard Grave, Piotr Bojanowski, and Tomas Mikolov. 2017.
\newblock Bag of tricks for efficient text classification.
\newblock In \emph{EACL: Volume 2, Short Papers}, pages 427--431, Valencia,
  Spain. ACL.

\bibitem[{Kim(2014)}]{kim2014}
Yoon Kim. 2014.
\newblock Convolutional neural networks for sentence classification.
\newblock In \emph{Proceedings of the 2014 Conference on Empirical Methods in
  Natural Language Processing ({EMNLP})}, pages 1746--1751, Doha, Qatar. ACL.

\bibitem[{Le and Mikolov(2014)}]{pmlrle14}
Quoc Le and Tomas Mikolov. 2014.
\newblock Distributed representations of sentences and documents.
\newblock In \emph{Proceedings of the 31st International Conference on Machine
  Learning}, volume~32 of \emph{Proceedings of Machine Learning Research},
  pages 1188--1196, Bejing, China. PMLR.

\bibitem[{Li and Roth(2002)}]{li2002}
Xin Li and Dan Roth. 2002.
\newblock Learning question classifiers.
\newblock In \emph{{COLING}}.

\bibitem[{Liu et~al.(2016)Liu, Qiu, and Huang}]{Liu2016}
Pengfei Liu, Xipeng Qiu, and Xuanjing Huang. 2016.
\newblock Recurrent neural network for text classification with multi-task
  learning.
\newblock In \emph{IJCAI}, page 2873–2879.

\bibitem[{Liu et~al.(2020)Liu, Wang, Zhang, You, Wu, and
  Dou}]{Liu2020LabelguidedLF}
X.~Liu, Song Wang, X.~Zhang, Xinxin You, J.~Wu, and D.~Dou. 2020.
\newblock Label-guided learning for text classification.
\newblock \emph{ArXiv}, abs/2002.10772.

\bibitem[{Luo et~al.(2017)Luo, Uzuner, and Szolovits}]{Luo2017BridgingSA}
Yuan Luo, {\"O}zlem Uzuner, and Peter Szolovits. 2017.
\newblock Bridging semantics and syntax with graph algorithms -
  state-of-the-art of extracting biomedical relations.
\newblock \emph{Briefings in bioinformatics}, 18 1:160--178.

\bibitem[{Manning et~al.(2008)Manning, Raghavan, and
  Schütze}]{manning_raghavan_schutze_2008}
Christopher~D. Manning, Prabhakar Raghavan, and Hinrich Schütze. 2008.
\newblock \emph{Introduction to Information Retrieval}.
\newblock Cambridge University Press.

\bibitem[{Mikolov et~al.(2013)Mikolov, Sutskever, Chen, Corrado, and
  Dean}]{Mikolov2013DistributedRO}
Tomas Mikolov, Ilya Sutskever, Kai Chen, Greg~S Corrado, and Jeff Dean. 2013.
\newblock Distributed representations of words and phrases and their
  compositionality.
\newblock In \emph{NIPS, Nevada, USA}, volume~26, pages 3111--3119. Curran
  Associates, Inc.

\bibitem[{Pang and Lee(2005)}]{10.3115/1219840.1219855}
Bo~Pang and Lillian Lee. 2005.
\newblock Seeing stars: Exploiting class relationships for sentiment
  categorization with respect to rating scales.
\newblock In \emph{ACL}, page 115–124, Michigan, USA. ACL.

\bibitem[{Pennington et~al.(2014)Pennington, Socher, and
  Manning}]{Pennington2014GloVeGV}
Jeffrey Pennington, Richard Socher, and Christopher~D. Manning. 2014.
\newblock Glove: Global vectors for word representation.
\newblock In \emph{EMNLP, Doha, Qatar}, page 1532–1543.

\bibitem[{Qin et~al.(2020)Qin, Hu, and Liu}]{qin-etal-2020-feature}
Qi~Qin, Wenpeng Hu, and Bing Liu. 2020.
\newblock Feature projection for improved text classification.
\newblock In \emph{ACL}, pages 8161--8171, Online. ACL.

\bibitem[{Rousseau et~al.(2015)Rousseau, Kiagias, and
  Vazirgiannis}]{rousseau2015}
Fran{\c{c}}ois Rousseau, Emmanouil Kiagias, and Michalis Vazirgiannis. 2015.
\newblock Text categorization as a graph classification problem.
\newblock In \emph{ACL (Volume 1: Long Papers)}, pages 1702--1712, Beijing,
  China. ACL.

\bibitem[{Saha et~al.(2020)Saha, Granmo, and Goodwin}]{rupsa2020sentiment}
Rupsa Saha, Ole-Christoffer Granmo, and Morten Goodwin. 2020.
\newblock Mining interpretable rules for sentiment and semantic relation
  analysis using tsetlin machines.
\newblock In \emph{Artificial Intelligence XXXVII}, pages 67--78, Cham.
  Springer International Publishing.

\bibitem[{Serrano and Smith(2019)}]{serrano-attention}
Sofia Serrano and Noah~A. Smith. 2019.
\newblock Is attention interpretable?
\newblock In \emph{ACL}, pages 2931--2951, Florence, Italy. ACL.

\bibitem[{Shen et~al.(2018{\natexlab{a}})Shen, Wang, Wang, Min, Su, Zhang, Li,
  Henao, and Carin}]{shen2018}
Dinghan Shen, Guoyin Wang, Wenlin Wang, Martin~Renqiang Min, Qinliang Su, Yizhe
  Zhang, Chunyuan Li, Ricardo Henao, and Lawrence Carin. 2018{\natexlab{a}}.
\newblock Baseline needs more love: On simple word-embedding-based models and
  associated pooling mechanisms.
\newblock In \emph{ACL Volume 1: Long Papers}, pages 440--450, Melbourne,
  Australia. ACL.

\bibitem[{Shen et~al.(2018{\natexlab{b}})Shen, Wang, Wang, Min, Su, Zhang, Li,
  Henao, and Carin}]{shen2018baseline}
Dinghan Shen, Guoyin Wang, Wenlin Wang, Martin~Renqiang Min, Qinliang Su, Yizhe
  Zhang, Chunyuan Li, Ricardo Henao, and Lawrence Carin. 2018{\natexlab{b}}.
\newblock Baseline needs more love: On simple word-embedding-based models and
  associated pooling mechanisms.
\newblock In \emph{ACL (Volume 1: Long Papers)}, pages 440--450, Melbourne,
  Australia. ACL.

\bibitem[{Socher et~al.(2013)Socher, Bauer, Manning, and Ng}]{socher}
Richard Socher, John Bauer, Christopher~D. Manning, and Andrew~Y. Ng. 2013.
\newblock Parsing with compositional vector grammars.
\newblock In \emph{Proceedings of the 51st Annual Meeting of the Association
  for Computational Linguistics (Volume 1: Long Papers)}, pages 455--465,
  Sofia, Bulgaria. Association for Computational Linguistics.

\bibitem[{Tang et~al.(2015)Tang, Qu, and Mei}]{jianPTE}
Jian Tang, Meng Qu, and Qiaozhu Mei. 2015.
\newblock Pte: Predictive text embedding through large-scale heterogeneous text
  networks.
\newblock In \emph{Proceedings of the 21th ACM SIGKDD International Conference
  on Knowledge Discovery and Data Mining}, KDD '15, page 1165–1174, Sydney,
  NSW, Australia. Association for Computing Machinery.

\bibitem[{Tang et~al.(2020)Tang, Blincoe, and
  Kempa-Liehr}]{Tang2020EnrichingFE}
Yichen Tang, Kelly Blincoe, and A.~Kempa-Liehr. 2020.
\newblock Enriching feature engineering for short text samples by language time
  series analysis.
\newblock \emph{EPJ Data Science}, 9:1--59.

\bibitem[{Turian et~al.(2010)Turian, Ratinov, and Bengio}]{turian}
Joseph Turian, Lev-Arie Ratinov, and Yoshua Bengio. 2010.
\newblock Word representations: A simple and general method for semi-supervised
  learning.
\newblock In \emph{Proceedings of the 48th Annual Meeting of the Association
  for Computational Linguistics}, pages 384--394, Uppsala, Sweden. Association
  for Computational Linguistics.

\bibitem[{Wang and Manning(2012)}]{wang2012}
Sida Wang and Christopher Manning. 2012.
\newblock Baselines and bigrams: Simple, good sentiment and topic
  classification.
\newblock In \emph{ACL (Volume 2: Short Papers)}, pages 90--94, Jeju Island,
  Korea.

\bibitem[{{Yadav} et~al.(2021){Yadav}, {Jiao}, {Granmo}, and
  {Goodwin}}]{yadav2021sentiment}
Rohan~Kumar {Yadav}, Lei {Jiao}, Ole-Christoffer {Granmo}, and Morten
  {Goodwin}. 2021.
\newblock Human-level interpretable learning for aspect-based sentiment
  analysis.
\newblock In \emph{The Thirty-Fifth AAAI Conference on Artificial Intelligence
  (AAAI-21)}. AAAI.

\bibitem[{Yadav. et~al.(2021)Yadav., Jiao., Granmo., and
  Goodwin.}]{icaart21rohan}
Rohan~Kumar Yadav., Lei Jiao., Ole{-}Christoffer Granmo., and Morten Goodwin.
  2021.
\newblock Interpretability in word sense disambiguation using tsetlin machine.
\newblock In \emph{Proceedings of the 13th International Conference on Agents
  and Artificial Intelligence - Volume 2: ICAART,}, pages 402--409. INSTICC,
  SciTePress.

\bibitem[{Yang et~al.(2019)Yang, Dai, Yang, Carbonell, Salakhutdinov, and
  Le}]{NEURIPS2019_dc6a7e65}
Zhilin Yang, Zihang Dai, Yiming Yang, Jaime Carbonell, Russ~R Salakhutdinov,
  and Quoc~V Le. 2019.
\newblock Xlnet: Generalized autoregressive pretraining for language
  understanding.
\newblock In \emph{Advances in Neural Information Processing Systems},
  volume~32. Curran Associates, Inc.

\bibitem[{Zhu and Koniusz(2021)}]{zhu2021simple}
Hao Zhu and Piotr Koniusz. 2021.
\newblock Simple spectral graph convolution.
\newblock In \emph{International Conference on Learning Representations}.

\end{thebibliography}
\bibliographystyle{acl_natbib}

\end{document}